\documentclass[letterpaper, 10 pt, conference]{ieeeconf} 

\IEEEoverridecommandlockouts                             
\pdfoutput=1

\usepackage{lipsum}
\usepackage{amsmath}
\usepackage[table]{xcolor}
\usepackage{multirow}
\usepackage{booktabs}
\usepackage{siunitx}
\usepackage{threeparttable}
\usepackage{graphicx}

\title{\LARGE \bf
The Effect of Haptic Guidance during Robotic-assisted Motor Training is Modulated by Personality Traits
}

\author{Alberto Garz\'as-Villar$^{1,*}$, Caspar Boersma$^{1,*}$, Alexis Derumigny$^{2}$, Arkady Zgonnikov$^{1}$,\\ and Laura Marchal-Crespo$^{1,3}$
\thanks{This work was supported by the Health and Technology Convergence Alliance of TU Delft, Erasmus MC, and Erasmus University Rotterdam.}
\thanks{$^{*}$ These authors contributed equally to this work.
        }%
\thanks{$^{1}$A. Garz\'as-Villar, C. Boersma, A. Zgonnikov, and L. Marchal-Crespo are with the Department of Cognitive Robotics, Delft University of Technology, Delft, The Netherlands.
        }%
\thanks{$^{2}$A. Derumigny is with the Department of Applied Mathematics, Delft University of Technology, Delft, The Netherlands.
        }%
\thanks{$^{3}$L. Marchal-Crespo is with the Department of Rehabilitation Medicine, Erasmus MC, University Medical Center Rotterdam, Rotterdam, The Netherlands.
        {\tt\small Corresponding author: a.garzasvillar@tudelft.nl
        }}%
}

\begin{document}

\maketitle
\thispagestyle{empty}
\pagestyle{empty}

\begin{abstract}
The provision of robotic assistance during motor training has proven to be effective in enhancing motor learning in some healthy trainee groups as well as patients. Personalizing such robotic assistance can help further improve motor (re)learning outcomes and cater better to the trainee's needs and desires. However, the development of personalized haptic assistance is hindered by the lack of understanding of the link between the trainee's personality and the effects of haptic guidance during human-robot interaction.
To address this gap, we ran an experiment with 42 healthy participants who trained with a robotic device
to control a virtual pendulum to hit incoming targets either
with or without haptic guidance.
We found that certain personal traits affected how users adapt and interact with the guidance during training.
In particular, those participants with an ``Achiever gaming style'' performed better and applied lower interaction forces to the robotic device than the average participant as the training progressed.
Conversely, participants with the ``Free spirit game style'' increased the interaction force in the course of training.
We also found an interaction between some personal characteristics and haptic guidance. Specifically, participants with a higher ``Transformation of challenge'' trait exhibited poorer performance during training while receiving haptic guidance compared to an average participant receiving haptic guidance.
Furthermore, individuals with an external Locus of Control tended to increase their interaction force with the device, deviating from the pattern observed in an average participant under the same guidance.
These findings suggest that individual characteristics may play a crucial role in the effectiveness of haptic guidance training strategies.

\end{abstract}

\section{Introduction}

The acquisition of new motor skills (i.e., motor learning) is a fundamental part of our lives. From learning to walk to manipulating objects with complex dynamics, we never stop learning. A growing part of the population will also face situations involving the re-acquisition of lost functions \cite{Berghuis2015}, e.g., after a brain injury \cite{Raghavan2015}. Motor (re)learning is usually a long process that entails a complex interplay between high-intensity training and psychological factors such as motivation and attention \cite{Wulf2016}.

Robotic systems emerged as a promising avenue to support high-intensity training of motor skills and rehabilitation after neurological trauma \cite{Basalp2021}.
A commonly studied robotic-assisted training method is haptic guidance, i.e., the robot physically supports the trainees in achieving the desired movements. In line with the Challenge Point framework \cite{Guadagnoli2004}, haptic guidance has proven to be effective in enhancing motor learning, especially in people with an initially low skill level and in early-stage stroke patients \cite{Dehem2019, Basalp2021}. 
The general promises of haptic guidance for clinical applications have stimulated research into more personalized robotic assistance \cite{Basalp2021, Beckers2022}.
The online adaptation to patients' abilities \cite{Miguel-Fernandez2023} and psychological states, such as motivation and engagement \cite{Koenig2011}, has been shown to improve rehabilitation outcomes.
However, there is limited research on personalizing the training based on personal characteristics (e.g., \cite{Novak2014}), despite literature indicating that personality traits can influence psychological states and, consequently, impact motor learning \cite{Anderson2022, Chen2000, Parks2009, Kanfer1999}.
Personality traits reflect people's characteristics, such as patterns of thoughts, feelings, and behaviors that are consistent and stable over long periods \cite{Diener2019}. Utilizing these data might allow anticipating how likely an individual is to enter particular psychological states or engage with an assistive device.

Previous research attempts have sought connections between personal traits, states, robotic-assisted training, and motor learning. In sports, for instance, \textbf{autotelic personality}---defined by Csikszentmihalyi as the disposition to actively seek challenges and flow experiences \cite{Csikszentmihalhi2020}--- has shown potential as a predictor of engagement \cite{Mikicin2013}. Two subscales from the well-established autotelic personality questionnaire, namely the ``Transformation of challenge'' and ``Transformation of boredom,'' offer insights into how individuals are likely to transform challenge and boredom into engagement \cite{Tse2020}. This concept aligns with the Challenge Point framework \cite{Guadagnoli2004} and the Flow theory \cite{Csikszentmihalhi2020}, which highlight the role of perceived difficulty in enhancing motor learning. 

Given the gamified nature of robotic-assisted training, insights from game theory might also be valuable to enhance robotic training. For example, Marczewski introduced the \textbf{Hexad gaming style classification} attending to different personality profiles---namely disruptor, free spirit, achiever, player, socializer, and philanthropist--- as a means to design games according to the users' profiles to improve engagement \cite{Marczewski2015}. The ``Achiever'' and the ``Free Spirit'' gaming styles might be of particular interest for robot-assisted training, as the motivational aspects that move them are mastery and autonomy, directly associated with the degree of interaction with the robot. Related to perceived autonomy is the \textbf{Locus of Control} (LOC), defined as the extent to which individuals perceive they are in control of their actions and outcomes \cite{Rotter1966}. This construct might be essential to understanding how trainees react to the robot's assistance.

Despite the circumstantial evidence in the literature pointing at potential effects that the trainees' personality traits may have on the effectiveness of robot-aided training \cite{Acharya2018}, these effects remain mainly unexplored. To address this gap, we performed an experiment with 42 healthy participants who trained with a robotic device to control a virtual pendulum to hit incoming targets either with or without haptic guidance.
We evaluated the effect of haptic guidance and personality traits and their interaction on task performance and human-robot interaction forces during training. We hypothesized that
participants with a high ``Transformation of challenge'' or ``Achiever gaming style'' may be more prompt to enhance task performance than those showing low levels of those traits.
Conversely, those with elevated ``Transformation of boredom'' or a ``Free spirit gaming style'' may prefer exploring the device possibilities, and thus, result in higher interaction forces with the robot. 
High interaction force is also expected if the guidance is perceived as challenging, particularly by those participants with a pronounced ``Transformation of challenge.''
Finally, we expected that participants with an internal LOC would increase the interaction force, as they may reject the external influence, while the contrary was expected in participants with an external LOC.

\section{Methods}

\subsection{Participants}
Forty-two healthy participants gave informed consent to participate in the study, which was approved by the TU Delft Human Research Ethics Committee (HREC). Data from one participant had to be discarded due to technical problems, and a second participant was considered an outlier due to the poor performance during the game. The remaining participants (n = 40, age = $27\pm6$yrs, 19 female, 21 male, 4 left-handed, 35 right-handed, 1 ambidextrous)
had no prior experience with the pendulum game.

\subsection{Experimental setup}
\begin{figure}[th]
    \centering
      \includegraphics[width= .9\columnwidth]{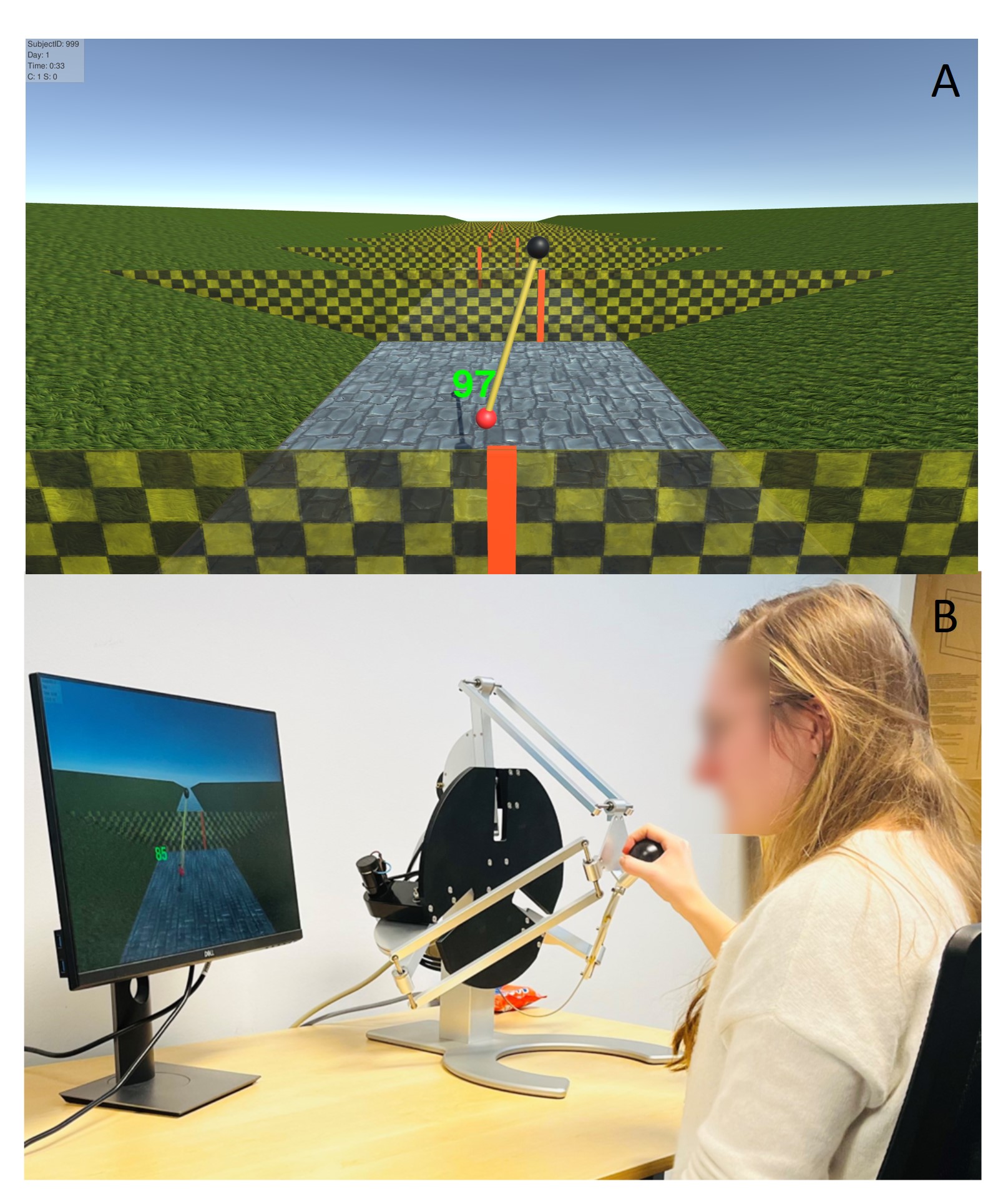}
    \caption{(\textbf{A}) Screenshot from the game representing the pendulum after hitting a target (vertical red line) and showing the score. (\textbf{B}) The experimental setup consisted of a Delta.3 haptic device (Force Dimension, Switzerland) and a display monitor where the pendulum game was presented.}
    \label{fig:Experimentalsetup}
\end{figure}

The experimental setup comprised a commercial robotic haptic device with three translational degrees of freedom (Delta.3, Force Dimension, Switzerland) integrated with a monitor for displaying the virtual environment (Fig. \ref{fig:Experimentalsetup}B). The virtual environment (Fig. \ref{fig:Experimentalsetup}A) was developed using Unity (Unity Technologies, US). The robot's motion control was adapted from the work of \"{O}zen et al. \cite{Ozen2021}. 
The integrated system captures user data at a frequency of 1.67\,\si{kHz}. 

\subsection{The pendulum game}\label{sec:dynamics}
\begin{figure}[thpb]
    \centering
    
    \includegraphics[scale=0.44]{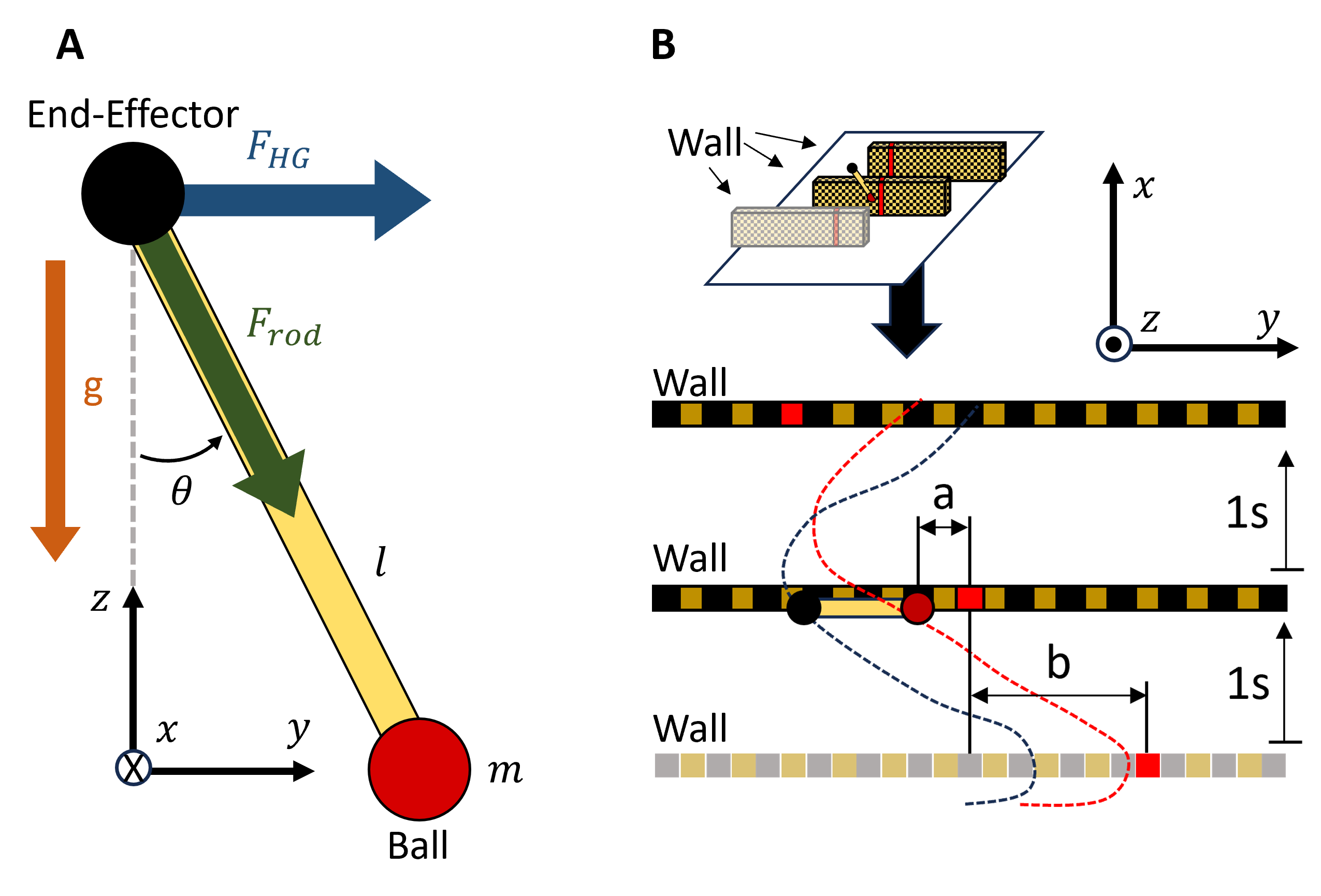}
    \caption{(\textbf{A})
    Schematic of the front view of the pendulum. The $F_{rod}$ is the force applied on the end-effector from the pendulum dynamics, while $F_{HG}$ represents the force from the haptic guidance. (\textbf{B}) Example of a top view of the game. The ball trajectory is represented with the red dashed line and the end-effector trajectory is shown with a black-dashed line. The horizontal lines represent the walls, and the red squares on those are the locations of the targets. The variable \textit{a} denotes the absolute error between the ball and the target positions on that wall. The variable \textit{b} is the position of the target w.r.t. to the game centerline. Adapted from \cite{Ozen2021}.}
    \label{fig:pendulum}
\end{figure}

We employed a similar pendulum game as the one described in detail in \cite{Ozen2021}. The game consisted of swinging a virtual pendulum by moving its pivoting point (black ball in Fig. \ref{fig:pendulum}A) through the manipulation of the robot end effector (1:1 movement mapping) to hit incoming targets with the pendulum mass (red ball in Fig. \ref{fig:pendulum}A). The virtual pendulum moved following the equation of motion of a simple pendulum, as implemented in \cite{Ozen2021}, with mass $m = 0.6$\,\si{kg} and rod length $l =  0.250$\,\si{m}. The movement of the pendulum was restricted to the frontal vertical plane (\textit{yz} plane in Fig. \ref{fig:pendulum}A). The pendulum dynamics, such as inertia and gravity, were reflected back to the robot end effector (haptic rendering, $F_{rod}$ in Fig. \ref{fig:pendulum}A) as described in \cite{Ozen2021}.

The targets were depicted as vertical red lines on semitransparent walls moving towards the participants (Fig. \ref{fig:Experimentalsetup}A). The targets were separated by one second and positioned on the walls at three different possible distances to the game centerline, i.e., either at the centerline or at $\pm$ 0.12\,\si{m} from the centerline (Fig. \ref{fig:pendulum}B). 
A total of 20 targets were presented per set. After passing a wall with the pendulum, a score was presented for 0.5\,\si{s} to provide feedback about the just performed ``hit'' (see  \cite{Ozen2021} for more details). A final score with the mean score across the 20 targets was also presented.

\subsection{Training strategies}
Participants were divided into two training groups: i) The \textit{Experimental} group, which trained with haptic guidance from the robot together with the haptic rendering from the pendulum dynamics; and ii) The \textit{Control} group, which trained without assistance but still with haptic rendering.  

The haptic guidance was designed to guide the participants to follow an optimal end-effector trajectory (in the \textit{y}-direction) between walls (or between the start position and the first wall) to maximize the performance in the target-hitting task. The reference trajectory ($y_{ref}$) was calculated every time the pendulum passed a wall, based on the pendulum's current states and the position of the next target, similar to \cite{Ozen2021}. The optimization was implemented using a quadratic function from the ACADO toolkit \cite{Houska2011}.
The cost function included the minimization of the distance between the foreseen pendulum ball and the target position on the next wall while maximizing the ball stabilization and minimizing the end-effector acceleration.

We selected a Proportional Derivative (PD)-controller to provide haptic guidance, as this is the most used controller in literature \cite{Basalp2021}. The PD-controller aimed to reduce the error $e(t)$ between the end-effector position $y_{EE}$ and the calculated reference trajectory $y_{ref}$ at each time step at the system frequency (1.67\,\si{kHz}). The proportional and derivative gains were set to 75.0\,\si{N/m} and 15.0\,\si{Ns/m} respectively. The maximum force that the Delta.3 robot can apply is 20.0\,\si{N}.

\begin{figure*}[t]
    \centering
   \includegraphics[width=.9\textwidth]{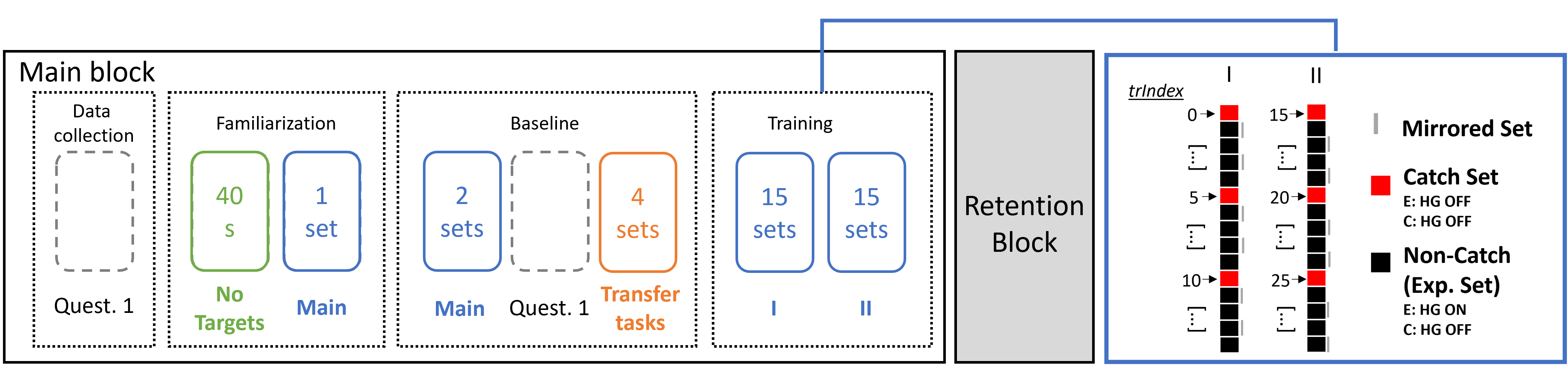}
    \caption{Study protocol. A set comprised of 20 targets. Quest: Questionnaire; Exp: Experimental; HG: Haptic Guidance.}
    \label{fig:protocol}
\end{figure*}

\subsection{Study protocol}

The overview of the study protocol is shown in Fig. \ref{fig:protocol}.
The experiment was performed in two different locations: Delft University of Technology, Delft, the Netherlands, and Alten Netherlands B.V., in Rotterdam, the Netherlands. 
Participants were allocated to either the \textit{Control} or \textit{Experimental} groups following an adaptive randomization method, i.e., the first 20 participants were randomly allocated, while the others were allocated based on their sex and questionnaire results to ensure even distribution between groups \cite{Basalp2021}. 

Participants sat in a comfortable chair with a backrest in front of a table where the screen and haptic device were located. The haptic device was placed at a comfortable reachable distance from the participant's dominant hand. The experiment started with the collection of demographic data and a battery of questionnaires to capture relevant personality traits (see section \ref{sec:Questionnaires}). Participants were then invited to familiarize themselves with the virtual environment and the haptic device for 40 seconds by moving the haptic end-effector and seeing the effect of their movements on the virtual pendulum. They could also feel the dynamics of the pendulum through the haptic rendering. They were then briefed about the pendulum game goal and were invited to perform a first set of 20 targets. 

After familiarization, the main experiment started. It consisted of three main phases: baseline, training, and retention. The haptic rendering was provided through all phases. During the baseline phase, the main task aimed to asses the initial skill level of the participants. This main task consisted of hitting 20 consecutive targets with the pendulum with dynamics described in Section \ref{sec:dynamics}. Also during this phase, two other similar tasks aimed to assess the generalization of the learned skill (out of the scope of this paper). 

After the baseline, participants trained in the main task with haptic guidance or no guidance, depending on the training group they were allocated to. The training phase was divided into two blocks of 15 sets each, and participants were allowed to rest between blocks. Participants in the \textit{Experimental} group did not receive haptic guidance on every fifth set without being informed (i.e., a total of six ``catch sets,''  Fig.~\ref{fig:protocol}) to prevent them from relying on the guidance during training. 
The position of the 20 targets per training set was the same as during the main task baseline in half of the sets, while the other half had the target positions located in the opposite value on the y-axis, i.e., were mirrored, to limit potential participants' boredom. 

Right after the last training set and 1--3 days later, participants performed a short- and long-term retention test, respectively, which followed the same structure as baseline.
In the following of this paper, we focus on the analysis of only the training phase. 

\subsection{Outcome metrics} \label{sec:outcome_metrics}

\subsubsection{Performance and human-robot interaction metrics}
The performance in the pendulum game was measured as the absolute difference ($|Error|$), in meters, between the red ball y-position at the moment of hitting the
wall and the target position. The human-robot interaction force ($|IntForce|$) was estimated with Reaction Torque Observers  as implemented by \"{O}zen et al. \cite{Ozen2021}.

\subsubsection{Personality traits questionnaires} \label{sec:Questionnaires}
Before the baseline phase, participants filled a battery of personality trait questionnaires; namely, the LOC scale \cite{Rotter1966}, the ``Achiever,'' and ``Free spirit'' sections of the Hexad Gaming style questionnaire \cite{Marczewski2015}, and the sub-questionnaires ``Transformation of challenge'' and ``Transformation of boredom'' 
from the Autotelic personality questionnaire \cite{Tse2020}. The questions from each questionnaire were averaged and normalized to range from 0 (low level of this trait/characteristic) and 1 (high level of this trait/characteristic), except the LOC, which was normalized to range from -1 (internal LOC) to 1 (external LOC).

\subsection{Statistical analysis}
We aimed to evaluate the relationship between personality traits and haptic guidance on task performance ($|Error|$) and human-robot interaction force ($|IntForce|$) during training. The data set was organized at a wall-target level, i.e., one data frame per target and participant  (0 $\leq$ \textit{wIndex} $\leq$ 19). An index (0 $\leq$ \textit{trIndex} $\leq$ 29) identified the order of the training sets. Participants were allocated an identifier number (ID).

We analyzed the data using two linear mixed models (LMM), 
with dependent variables the logarithmic scales of the metrics $|Error|$ and $|IntForce|$, respectively.
The logarithmic transformation was performed as the data was found to be skewed. 
We included in the model the outcomes from the ``Achiever gaming style'' (\textit{AC}), ``Free Spirit gaming style'' (\textit{FS}), ``Transformation of challenge'' (\textit{CH}), ``Transformation of boredom'' (\textit{BO}), and Locus of Control (\textit{LOC}) questionnaires.
The mean from each personality outcome was calculated and centered on the mean, denoted as  ``${XX}_{c}$''. 
The variable $Received$ $guidance$ (\textit{RG}) is categorical and \texttt{FALSE} if the guidance was not provided during the set, i.e., for all sets in the \textit{Control} group and catch trials in the \textit{Experimental} group. 
The participants' $ID$ and the index of the wall (\textit{wIndex}) were set as random effects. This decision was taken to incorporate dependencies of data from the same participant or wall, as well as approach random effects attributed to variability among different participants and target locations. 
The function \texttt{lmer} from the \texttt{lme4} package in \texttt{R} was used to fit the following models:
\begin{align*}
\log_{10}| \text{VAR}| = \,
&({AC}_{c} + {FS}_{c} + {LC}_{c} + {BD}_{c} + {LOC}) \, \times \\
&  ({RG} + {trIndex})+ (1|{ID}) + (1|{wIndex}),
\end{align*}
where the variable VAR was  $Error$ or $IntForce$ during the corresponding analysis. The variable describing the group (\textit{Experimental} or \textit{Control}) was excluded since our interest lay exclusively in the performance and interaction metrics with/without guidance, included in the variable \textit{RG}. 
We set significance to $p < 0.05$ and corrected for multiple comparisons with Bonferroni correction.

\begin{table*}[ht]
\centering
\scalebox{.97}{%
\begin{threeparttable}
\caption{Linear mixed model results for the performance metric \textit{$|Error|$} and human-robot interaction force \textit{$|IntForce|$}.}
\label{tab:table_results}
\begin{tabular}{lrrrrrrrr}
\toprule
\multirow{2}{*}{\textbf{Variable}} & \multicolumn{4}{c}{\textbf{$\log_{10}(|Error|)$}} & \multicolumn{4}{c}{\textbf{$\log_{10}(|IntForce|)$}} \\
\cmidrule(lr){2-5} \cmidrule(lr){6-9}
& \textbf{Estimate ($\beta$)} & \textbf{Std. Error} & \textbf{t value} & \textbf{Corr. p-value} & \textbf{Estimate ($\beta$)} & \textbf{Std. Error} & \textbf{t value} & \textbf{Corr. p-value}\\
\midrule
(Intercept) & -1.31 & 3.56e-02 & -36.80 & \textbf{$<$ 2e-16 ***} & -2.33e-01 & 2.45e-02& -9.48 & \textbf{1.24e-09 ***} \\
Achiever ($AC_{c}$) & 3.22e-01 & 2.65e-01 & 1.21 & 1 & 9.27e-02 & 1.41e-01 & 0.66 & 1\\
Free Spirit ($FS_{c}$) & -1.32e-01 & 3.80e-01 & -0.35 & 1 & -6.16e-02 & 2.02e-01 & -0.31 & 1\\
Transformation of challenge ($CH_{c}$) & -2.48e-01 & 3.00e-01 & -0.83 & 1 & -1.12e-01 & 1.60e-01 & -0.70 & 1\\
Transformation of boredom ($BO_{c}$) & 1.33e-01 & 1.59e-01 & 0.83 & 1 & -1.09e-02 & 8.47e-02 & -0.13 & 1\\
Locus of Control ($LOC$) & -2.32e-02 & 7.35e-02 & -0.32 & 1 & -1.49e-02 & 3.91e-02 & -0.38 & 1\\
Received guidance ($RG$) & -1.50e-01 & 1.08e-02 & -13.89 & \textbf{$<$ 2e-16 ***} & 7.22e-01 & 4.31e-03 & 167.49 & \textbf{$<$ 2e-16 ***}\\
Training index ($trIndex$) & -5.69e-03 & 3.61e-04 & -15.75 & \textbf{$<$ 2e-16 ***} & -1.56e-03 & 1.44e-04 & -10.85 & \textbf{$<$ 2e-16 ***}\\
$AC_{c}$ - $RG$ & -4.60e-02 & 1.19e-01 & -0.39 & 1 & -4.28e-03 & 4.76e-02 & -0.09 & 1\\
$AC_{c}$ - $trIndex$ & -1.80e-02 & 4.06e-03 & -4.44 & \textbf{1.61e-04 ***} & -1.17e-02 & 1.61e-03 & -7.27 & \textbf{6.49e-12 ***}\\
$FS_{c}$ - $RG$ & 4.09e-02 & 1.65e-01 & 0.25 & 1 & -1.04e-01 & 6.61e-02 & -1.57 & 1\\
$FS_{c}$ - $trIndex$ & 5.69e-04 & 5.80e-03 & 0.10 & 1 & 8.46e-03 & 2.30e-03 & 3.67 & \textbf{4.31e-03 **}\\
$CH_{c}$ - $RG$ & 4.34e-01 & 1.42e-01 & 3.06 & \textbf{4.01e-02 *} & 1.28e-01 & 5.69e-02 & 2.25 & 4.40e-01\\
$CH_{c}$ - $trIndex$ & 4.45e-03 & 4.59e-03 & 0.97 & 1 & 4.62e-03 & 1.82e-03 & 2.54 & 2.00e-01\\
$BO_{c}$ - $RG$ & -4.68e-02 & 7.33e-02 & -0.64 & 1 & 1.85e-02 & 2.93e-02 & 0.63 & 1\\
$BO_{c}$ - $trIndex$ & -3.94e-03 & 2.44e-03 & -1.62 & 1 & 4.73e-04 & 9.67e-04 & 0.49 & 1\\
$LOC$ - $RG$ & 3.86e-02 & 2.96e-02 & 1.30 & 1 & 6.69e-02 & 1.18e-02 & 5.66 & \textbf{2.77e-07 ***}\\
$LOC$ - $trIndex$ & 2.65e-03 & 1.12e-03 & 2.37 & 3.23e-01 & 2.03e-03 & 4.45e-04 & 4.57 & \textbf{8.80e-05 ***}\\
\bottomrule
\end{tabular}
\begin{tablenotes}
\item *($p < 0.05$), **($p < 0.01$), ***($p < 0.001$). AC: Achiever; FS: Free Spirit; CH: Transformation of challenge; BO: Transformation of boredom; LOC: Locus of Control; RG Received Guidance; trIndex: Training Index. 
Lowercase 'c' denotes centering the questionnaire results on the mean.
\end{tablenotes}
\end{threeparttable}%
}
\end{table*}

\section{Results}
The results of our analyses are summarized in Table I. 
When no guidance was applied, we did not find a main effect of the personality traits on the performance and interaction forces metrics. We found, however, a significant main effect of the training index on both behavioral metrics. In particular, participants' error decreased as training progressed (${\beta=-0.006}$, $t=-15.75$, $p < 2e-16$); for the average participant, the error was reduced by $1.30\%$ per training set.
After the first set, the participants also reduced the interaction forces by $0.36\%$, showing a 10\% reduction by the end of the 30 sets ($\beta=-1.56e-03$, $t=-10.85$, $p < 2e-16$).   
We also found main effects of the haptic guidance on the behavioral metrics.
When the haptic guidance was applied ($RG$ = \texttt{TRUE}), it reduced the error by $29\%$ for an average participant ($\beta= -0.015$, $t= -13.89$, $p < 2e-16$), and caused an increase of $400\%$ in the interaction forces ($\beta=7.22e-01$, $t=167.486$, $p < 2e-16$).

\subsection{Interaction effects of personality traits and training index}
The two traits related to gaming styles showed an interaction with the course of the training ($trIndex$) on the error and the interaction force in the absence of haptic guidance.
For example, for a participant with $0.1$ above-average ``Achiever gaming style''
the error was reduced by $1.71\%$ per set. Thus, at the end of the session, ``Achievers'' (0.1 over the average value) reduced the error by around $8\%$ more than the average participants. In the case of the interaction force, an increase of 0.1 in this trait resulted in a force reduction of 5.7\% w.r.t. to the average participant after the 30 sets. Furthermore, participants with 0.1 higher ``Free spirit'' increased the interaction force by 4.5\% after the training compared with the average participant.

\subsection{Interaction effects of personality and haptic guidance}
Haptic guidance was significantly less beneficial in reducing the error in participants with a high ``Transformation of challenge'' trait. For a participant with this trait 0.1 higher than the average, applying haptic guidance would only reduce the error by 21\% (compared to 29\% for the average participants). Still, this trait does not seem to affect the interaction force.

The LOC showed a significant interaction with the guidance on the force interaction metric. 
Participants with external LOC ($LOC = 1$) increased the interaction force by 12\%  w.r.t. the average participant receiving guidance, while participants with internal LOC ($LOC = -1$) decreased the interaction force by 11\%, deviating from the average participants receiving the same guidance.

\section{Discussion}
\subsection{Personality traits play a role in how participants adapt to the task through the duration of the training}
According to the motor learning literature, adaptation is expected to occur as training progresses \cite{Schmidt2018}. The observed error and force reduction through the training duration aligns with this expectation. Yet, not everybody seemed to follow the same tendency; as hypothesized, participants with an ``Achiever'' gaming style reduced even more those metrics at the end of the training session. This is in line with previous literature that showed that ``Achievers'' are motivated by mastery, and therefore, they look for self-improvement and challenges to overcome \cite{Marczewski2015}. Participants with this characteristic might have been more concerned about the task's main goal and aimed to maximize their performance and master the control of the pendulum by reducing the interaction force. However,
contrary to our hypothesis, no effect was found of the ``Transformation of boredom'' or ``Transformation of challenge'' traits on adaptation. A possible explanation for this unexpected finding is that the task might not have been perceived as too challenging or boring. Yet, since we did not measure participants' perceived task difficulty, this hypothesis needs further investigation.

Our findings also reveal that participants with a pronounced ``Free spirit gaming style'' increased the interaction force across the sets when the guidance was inactive. This diverges from the behavior of the average participant but aligns with our expectations and previous literature. Marczewski stated that players with a high ``Free spirit gaming style'' are motivated by exploration, encompassing the discovery of all potentialities within the device \cite{Marczewski2015}. It is conceivable that the inclination for exploration in participants with this characteristic extended to navigating the virtual environment and probing the limits of the haptic rendering, consequently exhibiting a higher interaction force.

\subsection{Personality traits play a role on how participants interact with the haptic guidance}
The primary goal of the haptic guidance was to reduce the error during the task by applying assisting forces. The significant error reduction and force increase associated with the guidance validate our approach and allow us to establish a reference for further comparisons.

We found an influence of several traits on how participants reacted to the haptic guidance. 
Individuals with a higher value of the ``Transformation of challenge'' trait ---characterized in the literature as feeling higher enjoyment of adverse situations \cite{Tse2020}--- exhibited a significantly higher error than the average participant when receiving guidance. This is contrary to our hypothesis as we expected these individuals to perceive the task goal as challenging, independently of whether guidance was provided, and therefore, prompting them to improve performance to overcome this difficulty. This unexpected result raises questions about how the participants' perceptions of the task's challenge evolved with the application of guidance.
The haptic guidance could have reduced the overall challenge of the task at hand to the point that they just lost interest. In addition, contrary to our hypothesis, no effect was found involving the ``Transformation of boredom'' trait when studying the impact of receiving haptic guidance on the error.
Results show very little effect of these ``Transformation'' traits on the interaction force, suggesting that further research should be conducted to explain how individuals perceive the difficulty of the task when haptic guidance is provided.
Together, our results show that while haptic guidance improves the performance metrics in our setup, some personality traits might limit its benefit in enhancing task performance during training. Other haptic strategies, e.g., error amplification \cite{Basalp2021}, might be more suited for people with a high ``Transformation of challenge'' characteristic.

Previous research indicated that participants with a more internal Locus of Control (LOC) tend to exhibit a higher percentage of interaction with the guidance algorithm and greater deviation from the guidance path during a robot-control task \cite{Acharya2018}. Thus, we hypothesized that participants with internal LOC would diverge more from the ideal path and, therefore, show higher force interaction when the guidance is applied compared to their counterparts. 
However, the contrary was found, i.e., higher interaction force among participants with external LOC. A plausible explanation lies in the nature of our haptic guidance approach, which recalculated the reference end-effector trajectory after each target based on the current states of the pendulum.
This type of guidance could have been perceived as more natural and less restrictive than the one employed in  \cite{Acharya2018}, which was based on obstacle avoidance. 
To gain a more comprehensive understanding, future research should explore the interplay between different haptic feedback delivery methods and individuals' Locus of Control.

\section{Conclusion}
We evaluated the relationship between the trainee's personality traits and haptic guidance during robot-assisted training of a virtual dynamic task. 
We found that certain personal traits affected how users adapt during the training in the absence of guidance. Compared to a participant with an average trait level, an increased ``Achiever gaming style'' led to improved performance and reduced interaction force along the training session.
In contrast, participants with a ``Free spirit gaming style'' increased the interaction force as training advanced, compared with the average participants.
We also found an interaction between some personality traits and haptic guidance. Remarkably, participants with a higher ``Transformation of challenge'' trait exhibited poorer performance during training while receiving haptic guidance, deviating from the average participant's performance under guidance. 
Furthermore, individuals with an external Locus of Control tended to increase their interaction force with the device compared to the average participant.
These findings suggest that individual characteristics may play a crucial role in the effectiveness of haptic guidance training strategies. 
These insights hold promise for personalizing robot-assisted training and neurorehabilitation strategies by tailoring interventions to individual characteristics. Yet, further research should be carried out to evaluate the effects of personality traits on motor learning and to find more suitable haptic strategies that might better suit these specific individuals.

\addtolength{\textheight}{-12cm}   

\bibliographystyle{IEEEtran}
\bibliography{IEEEabrv, bibliography}

\begin{thebibliography}{10}
\providecommand{\url}[1]{#1}
\csname url@samestyle\endcsname
\providecommand{\newblock}{\relax}
\providecommand{\bibinfo}[2]{#2}
\providecommand{\BIBentrySTDinterwordspacing}{\spaceskip=0pt\relax}
\providecommand{\BIBentryALTinterwordstretchfactor}{4}
\providecommand{\BIBentryALTinterwordspacing}{\spaceskip=\fontdimen2\font plus
\BIBentryALTinterwordstretchfactor\fontdimen3\font minus \fontdimen4\font\relax}
\providecommand{\BIBforeignlanguage}[2]{{%
\expandafter\ifx\csname l@#1\endcsname\relax
\typeout{** WARNING: IEEEtran.bst: No hyphenation pattern has been}%
\typeout{** loaded for the language `#1'. Using the pattern for}%
\typeout{** the default language instead.}%
\else
\language=\csname l@#1\endcsname
\fi
#2}}
\providecommand{\BIBdecl}{\relax}
\BIBdecl

\bibitem{Berghuis2015}
K.~M.~M. Berghuis, M.~P. Veldman, S.~Solnik, G.~Koch, I.~Zijdewind, and T.~Hortobágyi, ``Neuronal mechanisms of motor learning and motor memory consolidation in healthy old adults,'' \emph{AGE}, vol.~37, p.~53, Jun 2015.

\bibitem{Raghavan2015}
P.~Raghavan, ``Upper limb motor impairment after stroke,'' \emph{Physical Medicine and Rehabilitation Clinics of North America}, vol.~26, pp. 599--610, Nov 2015.

\bibitem{Wulf2016}
G.~Wulf and R.~Lewthwaite, ``Optimizing performance through intrinsic motivation and attention for learning: The optimal theory of motor learning,'' \emph{Psychonomic bulletin \& review}, vol.~23, pp. 1382--1414, 2016.

\bibitem{Basalp2021}
E.~Basalp, P.~Wolf, and L.~Marchal-Crespo, ``Haptic training: Which types facilitate (re)learning of which motor task and for whom? answers by a review,'' \emph{IEEE Transactions on Haptics}, vol.~14, no.~4, pp. 722--739, 2021.

\bibitem{Guadagnoli2004}
M.~A. Guadagnoli and T.~D. Lee, ``Challenge point: a framework for conceptualizing the effects of various practice conditions in motor learning,'' \emph{Journal of motor behavior}, vol.~36, no.~2, pp. 212--224, 2004.

\bibitem{Dehem2019}
S.~Dehem, M.~Gilliaux, G.~Stoquart, C.~Detrembleur, G.~Jacquemin, S.~Palumbo, A.~Frederick, and T.~Lejeune, ``Effectiveness of upper-limb robotic-assisted therapy in the early rehabilitation phase after stroke: A single-blind, randomised, controlled trial,'' \emph{Annals of physical and rehabilitation medicine}, vol.~62, no.~5, pp. 313--320, 2019.

\bibitem{Beckers2022}
N.~Beckers and L.~Marchal-Crespo, ``The role of haptic interactions with robots for promoting motor learning,'' in \emph{Neurorehabilitation Technology}.\hskip 1em plus 0.5em minus 0.4em\relax Springer, 2022, pp. 247--261.

\bibitem{Miguel-Fernandez2023}
J.~de~Miguel-Fern{\'a}ndez, J.~Lobo-Prat, E.~Prinsen, J.~M. Font-Llagunes, and L.~Marchal-Crespo, ``Control strategies used in lower limb exoskeletons for gait rehabilitation after brain injury: a systematic review and analysis of clinical effectiveness,'' \emph{Journal of neuroengineering and rehabilitation}, vol.~20, no.~1, p.~23, 2023.

\bibitem{Koenig2011}
A.~Koenig, X.~Omlin, L.~Zimmerli, M.~Sapa, C.~Krewer, M.~Bolliger, F.~M{\"u}ller, and R.~Riener, ``Psychological state estimation from physiological recordings during robot-assisted gait rehabilitation,'' \emph{Journal of rehabilitation research and development}, vol.~48, no.~4, pp. 367--385, 2011.

\bibitem{Novak2014}
D.~Novak, A.~Nagle, U.~Keller, and R.~Riener, ``Increasing motivation in robot-aided arm rehabilitation with competitive and cooperative gameplay,'' \emph{Journal of neuroengineering and rehabilitation}, vol.~11, no.~1, pp. 1--15, 2014.

\bibitem{Anderson2022}
D.~I. Anderson and A.~M. Williams, ``Individual differences in motor skill learning,'' \emph{Human movement science}, vol.~81, p. 102904, 2022.

\bibitem{Chen2000}
G.~Chen, S.~M. Gully, J.~A. Whiteman, and R.~N. Kilcullen, ``Examination of relationships among trait-like individual differences, state-like individual differences, and learning performance.'' \emph{Journal of applied psychology}, vol.~85, no.~6, p. 835, 2000.

\bibitem{Parks2009}
L.~Parks and R.~P. Guay, ``Personality, values, and motivation,'' \emph{Personality and individual differences}, vol.~47, no.~7, pp. 675--684, 2009.

\bibitem{Kanfer1999}
R.~Kanfer and E.~D. Heggestad, \emph{Individual differences in motivation: Traits and self-regulatory skills.}\hskip 1em plus 0.5em minus 0.4em\relax American Psychological Association, 1999, pp. 293--313.

\bibitem{Diener2019}
E.~Diener and R.~E. Lucas, ``Personality traits,'' \emph{General psychology: Required reading}, vol. 278, 2019.

\bibitem{Csikszentmihalhi2020}
M.~Csikszentmihalhi, \emph{Finding flow: The psychology of engagement with everyday life}.\hskip 1em plus 0.5em minus 0.4em\relax Hachette UK, 2020.

\bibitem{Mikicin2013}
M.~Mikicin, ``Autotelic personality as a predictor of engagement in sports,'' \emph{Biomedical Human Kinetics}, vol.~5, no.~1, pp. 84--92, 2013.

\bibitem{Tse2020}
D.~C.~K. Tse, V.~W. Lau, R.~Perlman, and M.~McLaughlin, ``The development and validation of the autotelic personality questionnaire,'' \emph{Journal of personality assessment}, vol. 102, no.~1, pp. 88--101, 2020.

\bibitem{Marczewski2015}
A.~Marczewski, ``Even ninja monkeys like to play,'' \emph{London: Blurb Inc}, vol.~1, no.~1, p.~28, 2015.

\bibitem{Rotter1966}
J.~B. Rotter, ``Generalized expectancies for internal versus external control of reinforcement.'' \emph{Psychological monographs: General and applied}, vol.~80, no.~1, p.~1, 1966.

\bibitem{Acharya2018}
U.~Acharya, S.~Kunde, L.~Hall, B.~A. Duncan, and J.~M. Bradley, ``Inference of user qualities in shared control,'' in \emph{2018 IEEE International Conference on Robotics and Automation (ICRA)}.\hskip 1em plus 0.5em minus 0.4em\relax IEEE, 2018, pp. 588--595.

\bibitem{Ozen2021}
{\"O}.~{\"O}zen, K.~A. Buetler, and L.~Marchal-Crespo, ``Promoting motor variability during robotic assistance enhances motor learning of dynamic tasks,'' \emph{Frontiers in neuroscience}, vol.~14, p. 600059, 2021.

\bibitem{Houska2011}
B.~Houska, H.~J. Ferreau, and M.~Diehl, ``Acado toolkit—an open-source framework for automatic control and dynamic optimization,'' \emph{Optimal Control Applications and Methods}, vol.~32, no.~3, pp. 298--312, 2011.

\bibitem{Schmidt2018}
R.~A. Schmidt, T.~D. Lee, C.~Winstein, G.~Wulf, and H.~N. Zelaznik, \emph{Motor control and learning: A behavioral emphasis}.\hskip 1em plus 0.5em minus 0.4em\relax Human kinetics, 2018.

\end{thebibliography}

\end{document}